\documentclass[letterpaper]{article} 
\usepackage{aaai2027} 
\usepackage[hyphens]{url} 
\usepackage{graphicx} 
\usepackage{natbib} 
\usepackage{caption} 
\usepackage{booktabs}
\usepackage{amsmath}
\usepackage{amssymb}
\usepackage{algorithm}
\usepackage{algorithmic}
\usepackage{xcolor}
\usepackage{colortbl}
\definecolor{highlightlight}{rgb}{0.9, 0.94, 0.98}

\title{Concentrate After Imagination: Text-Conditioned Evidence Grounding for Partially Relevant Video Retrieval}

\author{
    Shuaiqi Cheng\textsuperscript{\rm 1,2},
    Siyu You\textsuperscript{\rm 2},
    Yanbi Wu\textsuperscript{\rm 1},
    Yuxi Chen\textsuperscript{\rm 2},
    Jiahao Zhang\textsuperscript{\rm 1},
    Xuming Hu\textsuperscript{\rm 1,3}\thanks{Corresponding author.}
}
\affiliations{
    \textsuperscript{\rm 1}The Hong Kong University of Science and Technology (Guangzhou)\\
    \textsuperscript{\rm 2}University of Electronic Science and Technology of China\\
    \textsuperscript{\rm 3}The Hong Kong University of Science and Technology\\
    \texttt{scheng512@connect.hkust-gz.edu.cn, xuminghu97@gmail.com}
}

\nocopyright
\begin{document}

\maketitle

\begin{abstract}
Partially Relevant Video Retrieval (PRVR) retrieves untrimmed videos when queries describe only short moments. Although recent methods improve local representations, uncertainty modeling, and global context, final ranking often still trusts the strongest local response; a coincidentally similar fragment can therefore produce an unsupported peak. We identify this failure as the \textit{query-agnostic concentration bottleneck} and propose \textbf{TRACE}, a score-level evidence verification operator for PRVR. Given a query and global video registers, TRACE activates query-relevant registers, routes their support to frame-level evidence, and smoothly marginalizes alternative $q{\to}R{\to}F$ paths before localized temporal selection. Unlike representation-level feature fusion, TRACE uses this evidence only as a query-conditioned residual calibration of the original local score. On ActivityNet Captions, Charades-STA, and TVR, TRACE achieves the best SumR on all three benchmarks and improves the DreamPRVR backbone by $+1.2$, $+1.1$, and $+1.5$, respectively. Ablation, routing-corruption, hard-negative, and cross-backbone transfer analyses support the interpretation that the gains arise from query-conditioned evidence verification rather than a generic score offset.
\end{abstract}

\section{Introduction}

Text-to-Video Retrieval (T2VR) aims to retrieve videos relevant to a natural-language query from a large collection. Conventional T2VR benchmarks often assume that the query describes most or all of a short, trimmed clip~\cite{clip4clip,cap4video}. Real-world videos, however, are typically long and untrimmed, while a user may refer to only a brief event amid substantial irrelevant content. Partially Relevant Video Retrieval (PRVR)~\cite{ms-sl} addresses this setting: a video is considered relevant if it contains at least one temporal moment that matches the query.

PRVR exposes a tension between local concentration and reliable ranking. Because a query may match only a brief segment, the model must remain sensitive to local evidence; multiple-instance learning and max-style pooling over frame or clip similarities are therefore natural choices~\cite{mil_retrieval,ms-sl}. Yet the same mechanism makes ranking brittle: a negative video can be promoted by a coincidentally similar fragment. Reliable PRVR scoring must therefore answer two questions jointly: \textit{which local response matches the query}, and \textit{whether that response is corroborated by query-relevant global evidence from the video}.

\begin{figure*}[t]
    \centering
    \includegraphics[width=0.98\linewidth]{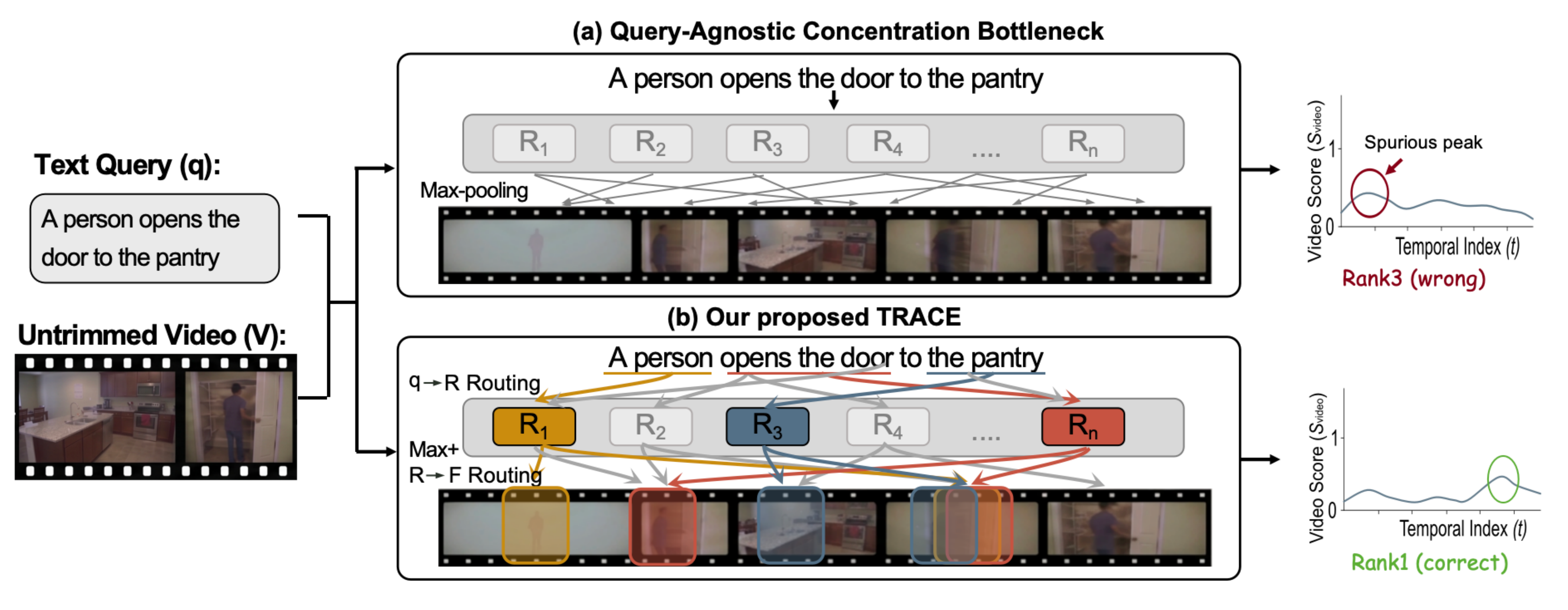}
    \caption{(a) Max-style concentration can select a spurious local peak because final scoring does not explicitly verify it with query-relevant global evidence, producing an incorrect rank. (b) TRACE activates relevant registers through $q{\to}R$ routing, grounds them back to temporal evidence through $R{\to}F$ routing, and concentrates only after this cross-granular support is formed. The illustrative score curves show the resulting correction from rank 3 to rank 1.}
   \label{fig:intro_motivation}
   \end{figure*}

Figure~\ref{fig:intro_motivation} contrasts direct local selection with TRACE's query-conditioned register activation and register-to-frame grounding. In the upper branch, a visually similar fragment can dominate even when it is unrelated to the query's global semantics; in the lower branch, only peaks supported by the activated registers are emphasized. The score curves are schematic rather than additional quantitative results: TRACE changes which peak is trusted while preserving localized temporal selection.

Recent PRVR research improves temporal modeling~\cite{ms-sl,gmmformer}, semantic alignment~\cite{sdm,hlformer}, uncertainty estimation~\cite{arl,holmes}, and global context~\cite{dreamprvr}. Yet richer representations do not guarantee reliable final concentration: the ranking function may still select a local peak without checking whether it is supported by query-relevant global context. We call this the \textit{query-agnostic concentration bottleneck}. Reliable scoring should preserve localized selection, select global evidence conditioned on the query, and ground that evidence back to temporal responses before concentration. Raw max pooling satisfies only the first requirement, while representation-level register fusion~\cite{dreamprvr} enriches visual tokens without explicitly verifying the winning response. TRACE therefore verifies a query-selected global-to-local path before concentration.

We propose \textbf{Text-Conditioned Register Activation and Cross-Granular Evidence Routing (TRACE)}, a hierarchical evidence concentration operator for PRVR. Query--register affinities activate relevant registers, register--frame compatibilities ground them temporally, and smooth log-sum-exp marginalization combines competing $q{\to}R{\to}F$ paths before localized maximization. The asymmetric design smooths alternative semantic explanations while retaining a hard temporal choice, preserving partial relevance while reducing unsupported peaks.

The distinction from ordinary feature fusion is at the ranking boundary. TRACE does not replace the backbone's learned representation or ask a new global embedding to dominate the score. Instead, it uses the query to decide which global anchors should support each candidate temporal response, and adds this verification signal through a small residual term. This makes the method compatible with register-based PRVR backbones while keeping the original local matching score as the primary retrieval signal. On ActivityNet Captions, Charades-STA, and TVR, TRACE achieves the highest SumR among compared methods, improving DreamPRVR~\cite{dreamprvr} by $+1.2$, $+1.1$, and $+1.5$. Ablation, routing-corruption, transfer, and hard-negative diagnostics support the verification-based explanation and distinguish the proposed routing from a generic score offset.

Our contributions are summarized as follows:
\begin{itemize}
 \item We identify and empirically diagnose the \textit{query-agnostic concentration bottleneck}: even when global context is encoded at the representation level, final concentration may still select a local response without query-conditioned global verification.
 \item We propose TRACE, a hierarchical evidence concentration operator that smoothly marginalizes text-conditioned global paths before localized maximum selection over register-grounded frame evidence.
 \item Across three PRVR benchmarks, TRACE yields consistent SumR improvements over strong baselines, while routing-corruption, cross-backbone transfer, and hard-negative diagnostics provide evidence that it is most effective on the intended unsupported-peak failure mode.
\end{itemize}

\section{Related Work}

\paragraph{Local Temporal Evidence Modeling in PRVR.}
PRVR methods model local temporal evidence to capture queries aligned with short segments. MS-SL~\cite{ms-sl} constructs multi-scale sliding-window representations; GMMFormer~\cite{gmmformer} and GMMFormerV2~\cite{gmmformerv2} refine temporal evidence through Gaussian-mixture modeling; DLDKD~\cite{dldkd} transfers knowledge from coarse to fine scales; and ProtoPRVR~\cite{protoprvr} reduces the candidate space using representative prototypes. Together, these approaches improve local evidence construction or search at the temporal-unit level.

\paragraph{Semantic Structure and Uncertainty in PRVR.}
Semantic alignment and uncertainty methods address ambiguous text--video relations and partial relevance. ARL~\cite{arl}, RAL~\cite{ral}, SDM~\cite{sdm}, and MSC-PRVR~\cite{mscprvr} improve semantic or uncertainty-aware modeling; HLFormer~\cite{hlformer} captures hierarchical partial relevance in hyperbolic space, while Holmes~\cite{holmes} combines evidential learning with optimal transport. These methods primarily improve supervision, representation geometry, or uncertainty estimation; they do not explicitly formulate the final local peak as a query-conditioned verification problem.

\paragraph{Global Context and Register-Based PRVR.}
Global-context methods complement local matching by injecting semantic priors through action-object modeling~\cite{a3prvr}, caption-driven alignment~\cite{captain}, contextual distillation~\cite{kdcnet}, and semantic-guided alignment~\cite{bcma}. Register tokens, first introduced in Vision Transformers for auxiliary global information~\cite{vit-register}, have also been adopted in multimodal and audio-visual models~\cite{falcon,regqav}. In PRVR, DreamPRVR~\cite{dreamprvr} generates diffusion-guided global registers and fuses them into frame and clip tokens. These methods mainly enrich representations before ranking, rather than directly calibrating the final score with query-conditioned register-grounded temporal evidence.

\paragraph{Evidence Aggregation in PRVR.}
Evidence aggregation converts local responses into a video-level retrieval score. Multiple-instance learning and max-pooling~\cite{mil_retrieval,ms-sl} remain standard; PRVR alternatives include Gaussian temporal aggregation~\cite{gmmformer}, prototype-based compression~\cite{protoprvr}, probabilistic relevance modeling~\cite{ral}, and optimal-transport assignment~\cite{holmes}. These designs primarily specify how local evidence is pooled, compressed, or assigned, whereas the separate question of whether the selected peak is supported by compatible global evidence is less explicitly studied. This separates \emph{evidence aggregation} from TRACE's cross-granular path-consistency verification. At a mathematical level, Gaussian components model temporal response distributions, while a transport plan aligns evidence under mass constraints. TRACE instead computes an unconstrained per-frame energy by log-sum-exp marginalization over $q{\to}R{\to}F$ paths, then uses it to test whether a local peak is globally supported before temporal max selection.

\section{Preliminaries}

\subsection{Problem Formulation}

Given a text query $Q$ and an untrimmed video $V$, Partially Relevant Video Retrieval (PRVR) ranks videos according to whether they contain a local temporal moment relevant to $Q$. The video branch produces frame embeddings $F=\{f_i\}_{i=1}^{M_f}\in\mathbb{R}^{M_f\times d}$ and clip embeddings $C=\{c_j\}_{j=1}^{M_c}\in\mathbb{R}^{M_c\times d}$, while the text branch produces a query embedding $q\in\mathbb{R}^{d}$. When a register-based backbone is used, the video branch additionally provides global registers $R=\{r_k\}_{k=1}^{N_r}\in\mathbb{R}^{N_r\times d}$.

For exposition, the backbone's original retrieval score can be written as a max-style combination over frame and clip responses:
\begin{equation}
S_{\mathrm{base}}(Q,V)=\alpha_f \max_i \cos(q,f_i)+\alpha_c \max_j \cos(q,c_j), \label{eq:base_score}
\end{equation}
where $\alpha_f+\alpha_c=1$ are the backbone's original frame--clip aggregation weights and are kept unchanged by TRACE. This maximum-based score fits partial relevance but concentrates on local responses without asking which global evidence is relevant to the current query or whether the selected local peak is supported by it.

\subsection{Register-Based PRVR}

DreamPRVR~\cite{dreamprvr} uses a coarse-to-fine register-generation pipeline: it first generates holistic video registers and then uses them to enhance fine-grained text--video matching, which can be summarized as the latent-register decomposition
\begin{equation}
p_{\theta,\phi}(Q|V)=\int \underbrace{p_{\theta}(Q|V,R)}_{\mathrm{matching}}\underbrace{p_{\phi}(R|V)}_{\mathrm{register\ generation}}dR, \label{eq:dreamprvr_pipeline}
\end{equation}
where $R=\{r_k\}_{k=1}^{N_r}$ are global registers generated from $V$. In practice, the generator produces register tokens that are fused with frame and clip embeddings through register-augmented attention, improving the representations used for retrieval. TRACE retains this register-based backbone and its representation-level context, while additionally using the registers as global evidence anchors selected by the current query at the scoring boundary. Thus, TRACE complements feature enhancement with score-level verification of the selected local peak.

\section{Method}
\vspace{-7pt}

\subsection{Overview}

Given query $q$, frame tokens $F$, clip tokens $C$, and registers $R$, TRACE constructs query-conditioned global-to-local evidence and integrates it with the base score as $S_{\mathrm{final}}=S_{\mathrm{base}}+\lambda S_{\mathrm{trace}}$. Here $q$ is produced by the text/query encoder, while $S_{\mathrm{base}}$ is the original DreamPRVR retrieval branch computed from the encoded local video tokens. TRACE adds a score-level residual, so $\lambda=0$ exactly recovers the backbone score. Figure~\ref{fig:trace_framework} summarizes the computation, with complete pseudocode provided in the Supplement Algorithm~S1. The following subsections detail the latent path view, cross-granular routing, hierarchical concentration, and learning objective.

\begin{figure*}[t]
 \centering
 \includegraphics[width=0.98\linewidth]{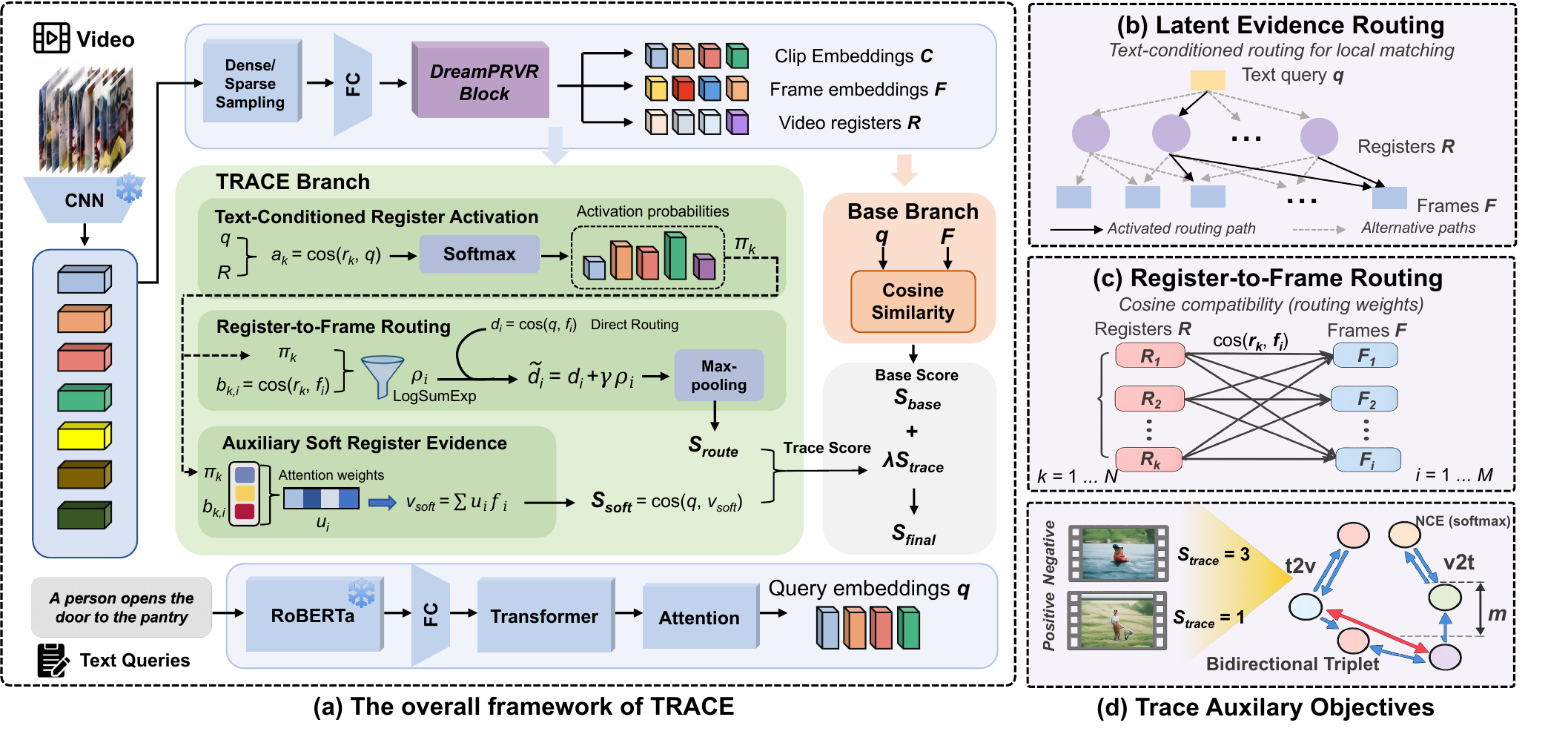}
 \caption{Overview of TRACE. (a) The overall framework: the video and text encoders produce clip embeddings $C$, frame embeddings $F$, registers $R$, and query embeddings $q$; the original base branch computes local matching over $C$ and $F$, while TRACE adds query-conditioned evidence through residual score fusion. (b) Latent $q{\to}R{\to}F$ routing paths, (c) register-to-frame compatibility, and (d) the auxiliary InfoNCE and bidirectional triplet objectives.}
 \label{fig:trace_framework}
\end{figure*}

\subsection{Text-Conditioned Latent Evidence Paths}
\label{sec:method-paths}

We view the missing scoring step as a text-conditioned latent path. Let $z$ denote a latent global evidence anchor selected by the query, and let $e$ denote a frame-level temporal evidence location. A schematic decomposition is given by
\begin{equation}
p_{\theta}(Q|V,R)\approx\sum_z p_{\theta}(z|Q,R)\sum_e p_{\theta}(e|z,V)p_{\theta}(Q|e,V), \label{eq:evidence_marginalization}
\end{equation}
where the expression is an energy-based scoring view rather than a calibrated likelihood. It separates query-conditioned anchor activation, anchor-to-frame grounding, and direct query--frame evidence. Raw max scoring mainly uses the last term, whereas TRACE combines all three before concentration.

\subsection{Cross-Granular Evidence Routing}
\label{sec:method-routing}

We instantiate the grounding principle with three score-level operations: direct query-to-frame matching, text-conditioned register activation, and register-to-frame routing. Before computing these scores, TRACE applies learned linear projections to the query, register keys, frame keys, and frame values, followed by $\ell_2$ normalization. For readability, we reuse $q$, $r_k$, and $f_i$ for the projected query, register key, and frame key, respectively, and denote the projected frame value by $f_i^v$.

\subsubsection{Query-to-Frame Direct Evidence.}
For each frame token $f_i$, we compute its direct compatibility with the query, $d_i=\cos(q,f_i)$, realizing $p_{\theta}(Q|e,V)$. By itself, this term can still be distracted by locally similar but globally unsupported frames.

\subsubsection{Text-Conditioned Register Activation.}
To select global semantic cues relevant to the current text, TRACE computes query-register affinity $a_k=\cos(q,r_k)$ and normalizes it as a softmax with temperature $\tau_r$:
\begin{equation}
\pi_k=\frac{\exp(a_k/\tau_r)}{\sum_{\ell=1}^{N_r}\exp(a_\ell/\tau_r)}, \label{eq:register_activation}
\end{equation}
approximating $p_{\theta}(z=k|Q,R)$.

\subsubsection{Register-to-Frame Routing.}
Each register is connected to frame-level evidence through register-frame compatibility $b_{k,i}=\cos(r_k,f_i)$. An expectation-style support for frame $i$ can be written as $\bar{g}_i=\sum_{k=1}^{N_r}\pi_k b_{k,i}$. Rather than using this diffuse average, TRACE aggregates the unnormalized joint path energies $a_k+b_{k,i}$ through
\begin{equation}
\rho_i = \tau_r \log \sum_{k=1}^{N_r} \exp\left((a_k+b_{k,i})/\tau_r\right). \label{eq:lse_route}
\end{equation}
This log-sum-exp form provides a differentiable soft maximum over latent $q{\to}R{\to}F$ paths. It gives high support only to frames compatible with query-relevant registers, so support varies with both query and temporal location. Register paths are smoothed, while temporal selection remains localized.

\subsection{Hierarchical Evidence Concentration}
\label{sec:method-concentration}

TRACE uses two asymmetric aggregation steps: smooth log-sum-exp over latent register paths (Eq.~\ref{eq:lse_route}), followed by hard temporal concentration. Multiple register paths may explain a query, but only a few temporal moments need to be selected.

\subsubsection{Routed Evidence Route.}
We combine direct evidence and register-routed support as
\begin{equation}
\tilde{d}_i=d_i+\gamma\rho_i, \label{eq:routed_frame_evidence}
\end{equation}
where $\gamma$ controls routing strength. Since PRVR relevance may occupy only a few moments, we select the strongest grounded response, $S_{\mathrm{route}}(Q,V)=\max_i \tilde{d}_i$. Thus the core operator performs smooth path marginalization followed by hard temporal selection, making unsupported negative peaks less likely to dominate.

\subsubsection{Soft Register Evidence Route.}
We include an auxiliary soft $q{\to}R{\to}F$ route to complement the hard temporal selector. Let
\begin{equation}
u_i = \sum_{k=1}^{N_r}\pi_k \frac{\exp(b_{k,i}/\tau_r)} {\sum_{j=1}^{M_f}\exp(b_{k,j}/\tau_r)}
\end{equation}
denote the query-conditioned register-to-frame attention. We form a soft evidence representation $v_{\mathrm{soft}}=\sum_i u_i f_i^v$ and score it as $S_{\mathrm{soft}}(Q,V)=\cos(q,v_{\mathrm{soft}})$. Unlike $S_{\mathrm{route}}$, which preserves sparse temporal concentration through a hard maximum, this branch aggregates frame values into a single soft representation before query scoring. It therefore serves as an auxiliary evidence branch that complements, rather than replaces, the core routed selector.

\subsubsection{Backbone Integration.}
The TRACE evidence score combines the core routed concentration score and the auxiliary soft evidence:
\begin{equation}
S_{\mathrm{trace}}(Q,V)=S_{\mathrm{route}}(Q,V)+\eta S_{\mathrm{soft}}(Q,V),
\end{equation}
where $\eta$ weights the auxiliary branch relative to the routed evidence. We then obtain the final retrieval score through residual fusion,
\begin{equation}
S_{\mathrm{final}}(Q,V)=S_{\mathrm{base}}(Q,V)+\lambda S_{\mathrm{trace}}(Q,V).
\end{equation}
The progressive schedule for introducing the evidence branch and ramping the effective fusion weight is described in the Implementation Details.

\subsection{Model Learning}

TRACE retains the backbone objective $\mathcal{L}_{\mathrm{base}}$, including retrieval supervision and register-generation regularization. Following the initial evidence-only warm-up, we jointly optimize the backbone and add $\mathcal{L}_{\mathrm{trace}}$ to make the evidence score discriminative before fusion.

For a mini-batch containing queries $\{Q_i\}_{i=1}^{B}$ and videos $\{V_j\}_{j=1}^{B_v}$, let $E_{ij}=S_{\mathrm{trace}}(Q_i,V_j)$ denote the TRACE evidence score and $y_i$ the positive-video index for $Q_i$. The evidence contrastive loss is the standard in-batch InfoNCE:
\begin{equation}
\mathcal{L}_{\mathrm{trace}}^{\mathrm{nce}}=-\frac{1}{B}\sum_{i=1}^{B}\log\frac{\exp(E_{i,y_i}/\tau_c)}{\sum_{j=1}^{B_v}\exp(E_{ij}/\tau_c)}, \label{eq:trace_nce}
\end{equation}
and the auxiliary objective additionally uses a bidirectional hinge-style triplet margin loss $\mathcal{L}_{\mathrm{trace}}^{\mathrm{tri}}=\mathcal{L}_{\mathrm{t2v}}^{\mathrm{tri}}+\mathcal{L}_{\mathrm{v2t}}^{\mathrm{tri}}$ over positive and negative query-video pairs with margin $m$. The TRACE auxiliary objective is $\mathcal{L}_{\mathrm{trace}} = \lambda_{\mathrm{nce}}\mathcal{L}_{\mathrm{trace}}^{\mathrm{nce}} + \lambda_{\mathrm{tri}}\mathcal{L}_{\mathrm{trace}}^{\mathrm{tri}}$, and the overall learning objective after introducing TRACE becomes $\mathcal{L}_{\mathrm{total}} = \mathcal{L}_{\mathrm{base}} + \mathcal{L}_{\mathrm{trace}}$.
The two evidence losses are complementary: InfoNCE separates positive videos from in-batch negatives, while the bidirectional triplet term enforces pairwise margins. $\mathcal{L}_{\mathrm{base}}$ preserves backbone retrieval and register generation, whereas $\mathcal{L}_{\mathrm{trace}}$ shapes the auxiliary evidence score. Fusion is delayed until the evidence branch becomes discriminative, preventing early errors from determining the ranking. The objective is introduced progressively in one continuous procedure: the evidence branch is warmed up without fusion, its auxiliary loss is then enabled, and fusion is ramped after stabilization. The exact schedule is given in the Implementation Details.

\section{Experiments}

\subsection{Experimental Setup}

\textbf{Datasets and Metrics.}
We evaluate ActivityNet Captions~\cite{activitynet_captions}, Charades-STA~\cite{charades_sta}, and TVR~\cite{xml} using standard PRVR splits~\cite{ms-sl}. We report Recall@$K$ for $K\in\{1,5,10,100\}$ and their sum, $\mathrm{SumR}=\mathrm{R@1}+\mathrm{R@5}+\mathrm{R@10}+\mathrm{R@100}$, in percentages.

\subsection{Implementation Details}

\textbf{Data Pre-Processing.} We use pre-extracted features throughout. ActivityNet Captions and Charades-STA use the I3D video and 1{,}024-dimensional RoBERTa text features released with MS-SL~\cite{ms-sl}. TVR uses 3{,}072-dimensional concatenated ResNet152/I3D video features and 768-dimensional RoBERTa text features.

\textbf{Experimental Configurations.} We adopt the DreamPRVR configuration~\cite{dreamprvr}: hidden dimension $384$, four attention heads, and $N_r=4,6,8$ registers for ActivityNet Captions, Charades-STA, and TVR. TRACE uses a unified configuration: $\lambda=0.03$, $\gamma=0.3$, $\eta=0.2$, and $\tau_r=0.07$. Training uses batch size $128$ on eight NVIDIA A800-80G GPUs. After a 100-epoch backbone warm-up, we freeze the backbone for the first five TRACE epochs and warm up the evidence projections without fusion; the auxiliary loss starts at $t=5$, and fusion starts at $t=30$ with a 20-epoch ramp. Full optimization details are in the Supplement.

\subsection{Comparison with State-of-the-art}

\subsubsection{Baselines.}
We compare conventional T2VR methods (CLIP4Clip~\cite{clip4clip}; Cap4Video~\cite{cap4video}), VCMR methods (XML~\cite{xml}; CONQUER~\cite{conquer}), and PRVR methods (MS-SL~\cite{ms-sl}, GMMFormer~\cite{gmmformer}, GMMFormerV2~\cite{gmmformerv2}, HLFormer~\cite{hlformer}, DreamPRVR~\cite{dreamprvr}, and Holmes~\cite{holmes}). DreamPRVR is the register-based backbone baseline. All results use the standard single-query protocol; for Holmes, we exclude its TVR dual-query variant because it requires additional inputs.

\subsubsection{Retrieval Performance.}
Table~\ref{tab:main} reports retrieval performance across the three benchmarks. PRVR-specific methods generally outperform conventional trimmed-video retrieval baselines, reflecting the value of partial localized modeling. TRACE achieves the highest SumR on all three datasets, exceeding the strongest prior SumR by $+0.5$, $+0.5$, and $+0.4$, and improving the DreamPRVR backbone by $+1.2$, $+1.1$, and $+1.5$ on ActivityNet Captions, Charades-STA, and TVR, respectively. The improvements appear at both ends of the ranking range: TRACE improves R@1 and R@100 on all three datasets, contributing to the overall SumR gain even when an intermediate cutoff, such as Charades-STA R@10, changes slightly. On TVR, TRACE also improves R@1 to $17.8\%$, suggesting that text-conditioned latent-path marginalization can refine final ranking decisions beyond representation-level register enhancement without replacing the underlying retrieval backbone.

\begin{table*}[t]
    \centering
    \small
    \setlength{\tabcolsep}{3pt}
    \begin{tabular}{lccccc|ccccc|ccccc}
    \toprule
    Method
    & \multicolumn{5}{c|}{ActivityNet}
    & \multicolumn{5}{c|}{Charades}
    & \multicolumn{5}{c}{TVR} \\
    & R@1 & R@5 & R@10 & R@100 & SumR
    & R@1 & R@5 & R@10 & R@100 & SumR
    & R@1 & R@5 & R@10 & R@100 & SumR \\
    \midrule
    CLIP4Clip & 5.9 & 19.3 & 30.4 & 71.6 & 127.3 & 1.8 & 6.5 & 10.9 & 44.2 & 63.4 & 9.9 & 24.3 & 34.3 & 72.5 & 141.0 \\
    Cap4Video & 6.3 & 20.4 & 30.9 & 72.6 & 130.2 & 1.9 & 6.7 & 11.3 & 45.0 & 65.0 & 10.3 & 26.4 & 36.8 & 74.0 & 147.5 \\
    XML & 5.3 & 19.4 & 30.6 & 73.1 & 128.4 & 1.6 & 6.0 & 10.1 & 46.9 & 64.6 & 10.7 & 28.1 & 38.1 & 80.3 & 157.1 \\
    CONQUER & 6.5 & 20.4 & 31.8 & 74.3 & 133.1 & 1.8 & 6.3 & 10.3 & 47.5 & 66.0 & 11.0 & 28.9 & 39.6 & 81.3 & 160.8 \\
    MS-SL & 7.1 & 22.5 & 34.7 & 75.8 & 140.1 & 1.8 & 7.1 & 11.8 & 47.7 & 68.4 & 13.5 & 32.1 & 43.4 & 83.4 & 172.4 \\
    GMMFormer & 8.3 & 24.9 & 36.7 & 76.1 & 146.0 & 2.1 & 7.8 & 12.5 & 50.6 & 72.9 & 13.9 & 33.3 & 44.5 & 84.9 & 176.6 \\
    GMMFormerV2 & 8.9 & 27.1 & 40.2 & 78.7 & 154.9 & 2.5 & 8.6 & 13.9 & 53.2 & 78.2 & 16.2 & 37.6 & 48.8 & 86.4 & 189.1 \\
    HLFormer & 8.7 & 27.1 & 40.1 & 79.0 & 154.9 & 2.6 & 8.5 & 13.7 & 54.0 & 78.7 & 15.7 & 37.1 & 48.5 & 86.4 & 187.7 \\
    DreamPRVR & 8.7 & 27.5 & 40.3 & 79.5 & 156.1 & 2.6 & 8.7 & 14.5 & 54.2 & 80.0 & 17.4 & 39.0 & 50.4 & 86.2 & 193.1 \\
    Holmes & \textbf{9.3} & 27.8 & 40.5 & 79.1 & 156.8 & 2.3 & \textbf{9.5} & \textbf{15.2} & 53.6 & 80.6 & 17.3 & 39.0 & 50.4 & \textbf{87.4} & 194.2 \\
    \rowcolor{highlightlight}\textbf{TRACE(Ours)} & 9.0 & \textbf{28.0} & \textbf{40.7} & \textbf{79.6} & \textbf{157.3} & \textbf{2.7} & 9.1 & 14.3 & \textbf{55.0} & \textbf{81.1} & \textbf{17.8} & \textbf{39.4} & \textbf{50.7} & 86.7 & \textbf{194.6} \\
    \bottomrule
    \end{tabular}
    \caption{Retrieval performance on ActivityNet Captions, Charades-STA, and TVR. \emph{R$@K$} denotes Recall@$K$ ($\%$), higher is better. SumR is computed from unrounded recall values. The best score in each column is marked in bold.}\vspace{-5pt}
    \label{tab:main}
    \end{table*}

The gain profile differs across benchmarks. ActivityNet improves across mid- and high-recall cutoffs, while Charades-STA shows stronger R@1 and R@100 effects with a small R@10 change. TVR obtains the largest SumR and R@1 gains over DreamPRVR, suggesting that query-conditioned support is especially useful when long videos contain several plausible hard negatives. TRACE therefore changes ranking selectively rather than uniformly increasing every recall cutoff.

\subsubsection{Qualitative Visualization.}
\begin{figure}[t]
 \centering
 \includegraphics[width=0.95\linewidth]{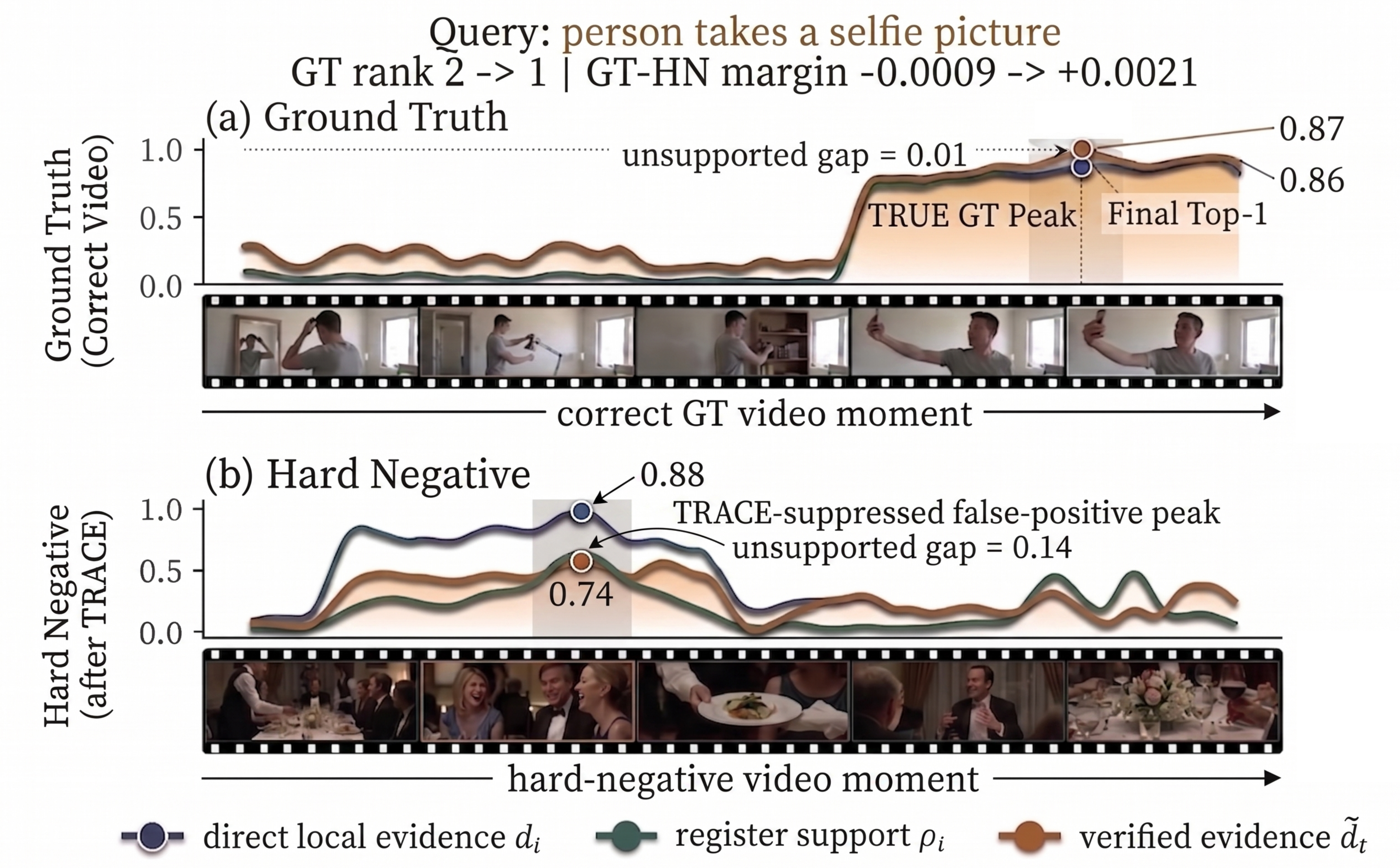}
\caption{Qualitative correction on Charades-STA. The top panel shows the ground-truth video and the bottom panel the baseline hard negative. TRACE restores the ground truth from rank 2 to rank 1.}\vspace{-5pt}
 \label{fig:trace_suppression_vs_gt}
\end{figure}

We select a Charades-STA case in which the register-based base branch ranks a hard negative above the ground-truth video. Figure~\ref{fig:trace_suppression_vs_gt} visualizes the corresponding score curves and video frames. The ground-truth moment receives aligned direct and register support, whereas the hard negative has a direct peak of $0.88$ but only $0.74$ register support, producing an unsupported gap of $0.14$. TRACE changes the ground-truth--hard-negative margin from $-0.0009$ to $+0.0021$ and restores the ground truth from rank 2 to rank 1.

\subsubsection{Model Efficiency.}
TRACE adds only lightweight query--register and register--frame score computations over a small number of registers. As shown in Table~\ref{tab:efficiency}, it improves retrieval performance with limited additional cost. We report per-epoch training time, parameter count, video feature extraction time over the full 1,334-video evaluation set, and retrieval time, including query encoding, similarity computation, and ranking. Relative to DreamPRVR, TRACE increases the parameter count by approximately $1.6\%$, feature extraction time by $2.9\%$, retrieval time by $0.6\%$, and per-epoch training time by $5.3\%$. This efficiency is consistent with our design: TRACE adds a lightweight score-time evidence selection rule without introducing or replacing a heavy video-language backbone.

\begin{table}[t]
\centering
\small
\setlength{\tabcolsep}{3.5pt} 
\resizebox{\linewidth}{!}{%
\begin{tabular}{lccccc}
\toprule
Model & \begin{tabular}[c]{@{}c@{}}Train Time\\ (ms/epoch)\end{tabular} & \begin{tabular}[c]{@{}c@{}}Params\\ (M)\end{tabular} & \begin{tabular}[c]{@{}c@{}}Infer Time\\ (ms)\end{tabular} & \begin{tabular}[c]{@{}c@{}}Retrieval\\ (ms)\end{tabular} & SumR \\
\midrule
GMMFormer & 26,887 & 12.85 & 2,876 & 3,238 & 72.9 \\
HLFormer & 31,463 & 28.43 & 3,816 & 3,655 & 78.7 \\
GMMFormerV2 & 38,004 & 30.79 & 3,843 & 3,688 & 78.2 \\
DreamPRVR & 33,609 & 36.14 & 4,001 & 3,686 & 80.0 \\
\textbf{TRACE} & 35,396 & 36.73 & 4,118 & 3,708 & \textbf{81.1} \\
\bottomrule
\end{tabular}%
}
\caption{Training and inference efficiency on Charades-STA. Inference time denotes feature extraction for 1,334 videos in the evaluation set, while retrieval time accounts for query encoding, similarity computation, and ranking.}\vspace{-5pt}
\label{tab:efficiency}
\end{table}

\subsection{Ablation Study}

\begin{table*}[t]
\centering
\small
\setlength{\tabcolsep}{2.5pt}
\begin{tabular}{llccccc|ccccc|ccccc}
\toprule
ID & Variant
& \multicolumn{5}{c|}{ActivityNet}
& \multicolumn{5}{c|}{Charades}
& \multicolumn{5}{c}{TVR} \\
& & R@1 & R@5 & R@10 & R@100 & SumR
& R@1 & R@5 & R@10 & R@100 & SumR
& R@1 & R@5 & R@10 & R@100 & SumR \\
\midrule
\rowcolor{highlightlight}(0) & \textbf{TRACE(full)} & 9.0 & 28.0 & 40.7 & 79.6 & 157.3 & 2.7 & 9.1 & 14.3 & 55.0 & 81.1 & 17.8 & 39.4 & 50.7 & 86.7 & 194.6 \\
\multicolumn{17}{l}{\emph{Efficacy of Evidence Source Construction}} \\
(1) & Base model & 8.7 & 27.5 & 40.3 & 79.5 & 156.1 & 2.6 & 8.7 & 14.5 & 54.2 & 80.0 & 17.4 & 39.0 & 50.4 & 86.2 & 193.1 \\
(2) & $S_{qF}$ only & 8.8 & 27.6 & 40.3 & 79.5 & 156.2 & 2.7 & 8.7 & 14.6 & 54.3 & 80.3 & 17.5 & 39.3 & 50.7 & 86.7 & 193.4 \\
(3) & $S_{\mathrm{soft}}$ only & 8.8 & 27.7 & 40.5 & 79.5 & 156.5 & 2.6 & 8.8 & 14.5 & 54.6 & 80.5 & 17.5 & 39.2 & 50.4 & 86.4 & 193.6 \\
(4) & $S_{qF}+\eta S_{\mathrm{soft}}$ & 8.9 & 27.8 & 40.6 & 79.6 & 156.9 & 2.7 & 9.0 & 14.4 & 54.8 & 80.9 & 17.6 & 39.1 & 50.8 & 86.8 & 194.3 \\
\multicolumn{17}{l}{\emph{Efficacy of Text-Conditioned Register Routing}} \\
(5) & $S_{qR}$ only & 8.8 & 27.7 & 40.5 & 79.5 & 156.5 & 2.6 & 8.8 & 14.4 & 54.6 & 80.4 & 17.5 & 39.2 & 50.5 & 86.5 & 193.8 \\
(6) & Uniform $q{\to}R$ & 8.9 & 27.7 & 40.6 & 79.6 & 156.6 & 2.6 & 9.0 & 14.3 & 55.0 & 80.8 & 17.6 & 39.2 & 50.6 & 86.7 & 194.0 \\
(7) & $S_{\mathrm{route}}$ only & 8.9 & 27.9 & 40.8 & 79.6 & 157.1 & 2.6 & 9.0 & 14.3 & 55.1 & 81.0 & 17.6 & 39.3 & 50.7 & 86.7 & 194.4 \\
\multicolumn{17}{l}{\emph{Efficacy of Evidence Aggregation}} \\
(8) & Avg$(\tilde{d}_i)$ & 8.8 & 27.7 & 40.5 & 79.5 & 156.5 & 2.7 & 8.8 & 14.5 & 54.4 & 80.4 & 17.7 & 39.1 & 50.5 & 86.5 & 193.8 \\
\bottomrule
\end{tabular}
\caption{Ablation and route diagnostics of TRACE. All evidence variants use the same integration weight $\lambda$; combined variants retain the same auxiliary weight $\eta$ as TRACE. Uniform $q{\to}R$ replaces query-conditioned register affinity with equal register weights, while Avg$(\tilde d_i)$ replaces the outer temporal maximum with a valid-frame average.}\vspace{-7pt}
\label{tab:ablation}
\end{table*}

We isolate three questions in Table~\ref{tab:ablation}: whether global evidence must be selected by the query, whether it must be grounded back to frames, and whether the grounded evidence should retain localized temporal concentration.

Rows (2) and (3) already improve over the base row (1) by using direct query--frame evidence $S_{qF}$ or register-guided soft evidence $S_{\mathrm{soft}}$; their combination in row (4) reaches 156.9 / 80.9 / 194.3 SumR, showing that the two signals are complementary. The most informative change is row (7), which replaces the direct-plus-soft aggregation with routed max evidence $S_{\mathrm{route}}$. Unlike the soft route, $S_{\mathrm{route}}$ preserves frame-wise path evidence and applies hard temporal concentration only after register-path marginalization. It improves over $S_{qF}$ alone by $+0.9$, $+0.7$, and $+1.0$ SumR on ActivityNet, Charades-STA, and TVR, respectively. Combining the routed selector with the auxiliary soft branch in row (0) gives the strongest overall result.

Rows (5), (6), and (8) further isolate the hierarchical operator. Row (5) drops frame grounding and underperforms, showing global matching alone is insufficient. Row (6) replaces query-conditioned activation with uniform register weights and falls behind row (7), providing evidence that latent paths should be selected by the current text. Row (8) replaces the hard temporal maximum with average aggregation, which dilutes short relevant moments. Together, these results provide evidence against three simpler explanations: a generic global offset, query-independent register weighting, or temporal averaging.

Compared with direct-plus-soft fusion, routed max evidence improves SumR by $+0.2$, $+0.1$, and $+0.1$ on the three datasets, while the auxiliary soft branch yields the final $157.3/81.1/194.6$. Although these margins are modest, their consistency, together with the corruption diagnostics, indicates that routing provides a distinct ranking signal.

\subsection{Register Activation Visualization}

\begin{figure}[t]
 \centering
 \includegraphics[width=0.95\linewidth]{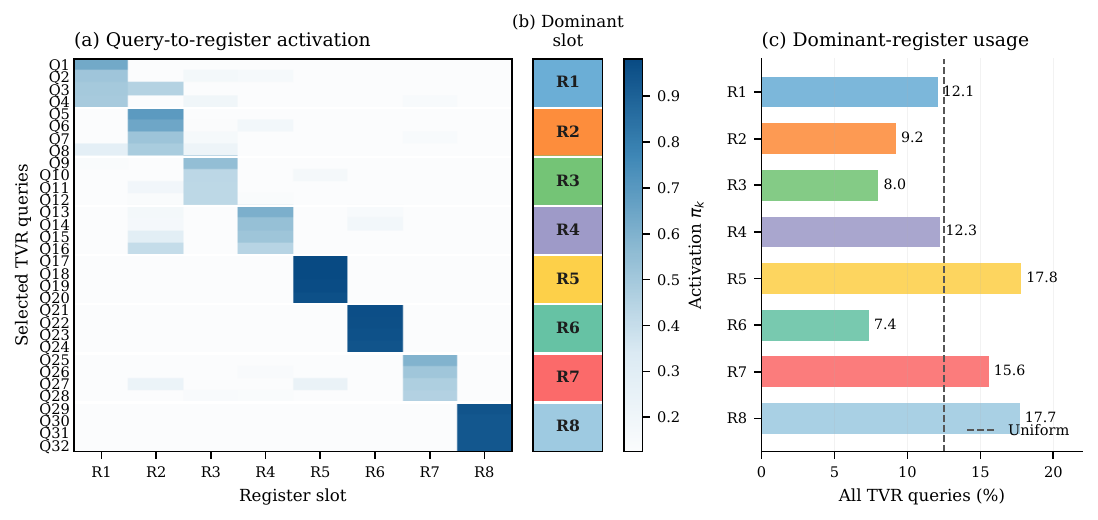}
 \caption{Visualization of text-conditioned register activation on TVR. (a) Query-to-register weights for selected pairs. (b) Dominant register per query group. (c) Dominant-register usage over all 10,895 TVR validation query-positive pairs. TRACE activates registers in a query-dependent manner.}\vspace{-7pt}
 \label{fig:register_activation}
\end{figure}

We visualize whether register activation varies with the query. Figure~\ref{fig:register_activation} shows query-to-register activation patterns on TVR. Panel (a) displays soft activation weights for representative query--video pairs, panel (b) summarizes the dominant register within query groups, and panel (c) reports aggregate usage over validation pairs. Different queries emphasize different register indices, while aggregate usage spans all indices rather than collapsing to one. This behavior complements Table~\ref{tab:ablation}: the uniform $q{\to}R$ variant in row (6) removes the query-dependent selection shown in the visualization.

\subsection{Cross-Backbone Transferability}
To test whether the proposed verification principle depends on diffusion-generated registers, we transfer the same score-time routing idea to four frozen, non-register-based PRVR backbones. For each backbone, lightweight global or temporal anchors and the fusion weight are selected on validation data and then frozen for evaluation. Table~\ref{tab:non_register_main} shows positive $\Delta$SumR on all three datasets for every backbone, with average gains ranging from $+0.17$ on MS-SL to $+0.96$ on HLFormer. These results suggest that learned registers are an effective instantiation of global evidence anchors, but the underlying verification rule is not restricted to a particular backbone.

\begin{table}[t]
\centering
\small
\setlength{\tabcolsep}{5pt}
\begin{tabular}{lcccc}
\toprule
Backbone & ActivityNet & Charades & TVR & Avg. \\
\midrule
MS-SL & +0.18 & +0.13 & +0.21 & +0.17 \\
GMMFormer & +0.24 & +0.47 & +0.50 & +0.40 \\
GMMFormerV2 & +0.54 & +0.31 & +0.88 & +0.58 \\
HLFormer & +0.89 & +0.56 & +1.43 & +0.96 \\
\bottomrule
\end{tabular}
\caption{Frozen-backbone diagnostic on representative earlier non-register-based PRVR backbones. We report evaluation-split $\Delta$SumR after selecting the anchor construction and fusion weight on validation data and freezing both choices. Full settings are in the Supplement Table~S1.}\vspace{-7pt}
\label{tab:non_register_main}
\end{table}

\section{Conclusion}

In this paper, we propose TRACE, a score-level evidence verification operator for partially relevant video retrieval. TRACE activates global registers according to the current query, grounds their support to frame-level evidence, and performs smooth latent-path aggregation before localized temporal selection. This design preserves the local concentration required by PRVR while reducing the influence of unsupported peaks, and can be integrated with an existing register-based backbone through residual fusion. Extensive experiments on three PRVR benchmarks show that TRACE consistently improves strong retrieval baselines, achieving SumR gains of $+1.2$, $+1.1$, and $+1.5$ over DreamPRVR, while ablation, routing-corruption, cross-backbone transfer, hard-negative, and qualitative analyses support the role of query-conditioned evidence grounding. TRACE offers a complementary perspective to representation-centric PRVR modeling: global context should not only enrich video representations, but also participate in verifying the evidence used for final ranking.

\bibliography{aaai2027}

\begin{thebibliography}{26}
\providecommand{\natexlab}[1]{#1}

\bibitem[{Chen et~al.(2026{\natexlab{a}})Chen, Zhou, Wang, Ning, Xiong, Li, Wen, and Tan}]{captain}
Chen, C.; Zhou, K.; Wang, F.; Ning, Y.; Xiong, Z.; Li, Y.; Wen, Z.; and Tan, M. 2026{\natexlab{a}}.
\newblock CaptAin: Caption-driven Alignment for Bridging Modality Gaps in Partially Relevant Video Retrieval.
\newblock In \emph{Proceedings of the IEEE/CVF Conference on Computer Vision and Pattern Recognition}, 6208--6217.

\bibitem[{Chen et~al.(2026{\natexlab{b}})Chen, Zhou, Wen, You, Li, Xiang, and Tan}]{a3prvr}
Chen, C.; Zhou, K.; Wen, Z.; You, Z.; Li, Y.; Xiang, T.; and Tan, M. 2026{\natexlab{b}}.
\newblock Action-and-object aware alignment for partially relevant video retrieval.
\newblock In \emph{Proceedings of the AAAI Conference on Artificial Intelligence}, volume~40, 2814--2822.

\bibitem[{Cho et~al.(2025)Cho, Moon, Jun, Jung, and Heo}]{arl}
Cho, C.-H.; Moon, W.; Jun, W.; Jung, M.; and Heo, J.-P. 2025.
\newblock Ambiguity-restrained text-video representation learning for partially relevant video retrieval.
\newblock In \emph{Proceedings of the AAAI Conference on Artificial Intelligence}, volume~39, 2500--2508.

\bibitem[{Darcet et~al.(2024)Darcet, Oquab, Mairal, and Bojanowski}]{vit-register}
Darcet, T.; Oquab, M.; Mairal, J.; and Bojanowski, P. 2024.
\newblock Vision transformers need registers.
\newblock In \emph{International conference on learning representations}, volume 2024, 2632--2652.

\bibitem[{Dietterich, Lathrop, and Lozano-P{\'e}rez(1997)}]{mil_retrieval}
Dietterich, T.~G.; Lathrop, R.~H.; and Lozano-P{\'e}rez, T. 1997.
\newblock Solving the multiple instance problem with axis-parallel rectangles.
\newblock \emph{Artificial intelligence}, 89(1-2): 31--71.

\bibitem[{Dong et~al.(2022)Dong, Chen, Zhang, Yang, Chen, Li, and Wang}]{ms-sl}
Dong, J.; Chen, X.; Zhang, M.; Yang, X.; Chen, S.; Li, X.; and Wang, X. 2022.
\newblock Partially relevant video retrieval.
\newblock In \emph{Proceedings of the 30th ACM International Conference on Multimedia}, 246--257.

\bibitem[{Dong et~al.(2023)Dong, Zhang, Zhang, Chen, Liu, Qu, Wang, and Liu}]{dldkd}
Dong, J.; Zhang, M.; Zhang, Z.; Chen, X.; Liu, D.; Qu, X.; Wang, X.; and Liu, B. 2023.
\newblock Dual learning with dynamic knowledge distillation for partially relevant video retrieval.
\newblock In \emph{Proceedings of the IEEE/CVF International Conference on Computer Vision}, 11302--11312.

\bibitem[{Gao et~al.(2017)Gao, Sun, Yang, and Nevatia}]{charades_sta}
Gao, J.; Sun, C.; Yang, Z.; and Nevatia, R. 2017.
\newblock Tall: Temporal activity localization via language query.
\newblock In \emph{Proceedings of the IEEE international conference on computer vision}, 5267--5275.

\bibitem[{Hou, Ngo, and Chan(2021)}]{conquer}
Hou, Z.; Ngo, C.-W.; and Chan, W.~K. 2021.
\newblock Conquer: Contextual query-aware ranking for video corpus moment retrieval.
\newblock In \emph{Proceedings of the 29th ACM International Conference on Multimedia}, 3900--3908.

\bibitem[{Jun et~al.(2025)Jun, Moon, Cho, Jung, and Heo}]{sdm}
Jun, W.; Moon, W.; Cho, C.-H.; Jung, M.; and Heo, J.-P. 2025.
\newblock Bridging the semantic granularity gap between text and frame representations for partially relevant video retrieval.
\newblock In \emph{Proceedings of the AAAI Conference on Artificial Intelligence}, volume~39, 4166--4174.

\bibitem[{Krishna et~al.(2017)Krishna, Hata, Ren, Fei-Fei, and Niebles}]{activitynet_captions}
Krishna, R.; Hata, K.; Ren, F.; Fei-Fei, L.; and Niebles, J.~C. 2017.
\newblock Dense-Captioning Events in Videos.
\newblock In \emph{2017 IEEE International Conference on Computer Vision (ICCV)}, 706--715. IEEE.

\bibitem[{Lei et~al.(2020)Lei, Yu, Berg, and Bansal}]{xml}
Lei, J.; Yu, L.; Berg, T.~L.; and Bansal, M. 2020.
\newblock Tvr: A large-scale dataset for video-subtitle moment retrieval.
\newblock In \emph{European Conference on Computer Vision}, 447--463. Springer.

\bibitem[{Li et~al.(2026{\natexlab{a}})Li, Zhao, Zhang, and Wen}]{bcma}
Li, H.; Zhao, J.; Zhang, Y.; and Wen, J. 2026{\natexlab{a}}.
\newblock Bidirectional Cross-Modal Collaborative Alignment via Semantic-Guided Visual Embeddings for Partially Relevant Video Retrieval.
\newblock \emph{IEEE Transactions on Image Processing}.

\bibitem[{Li et~al.(2026{\natexlab{b}})Li, Lai, Lou, Wang, Wang, Chen, Wang, and Xia}]{holmes}
Li, J.; Lai, P.; Lou, X.; Wang, J.; Wang, Y.; Chen, K.; Wang, Y.; and Xia, S.-T. 2026{\natexlab{b}}.
\newblock Revisiting Uncertainty: On Evidential Learning for Partially Relevant Video Retrieval.
\newblock In \emph{Forty-third International Conference on Machine Learning}.

\bibitem[{Li et~al.(2026{\natexlab{c}})Li, Lou, Wang, Wang, Wang, Xia, and Chen}]{dreamprvr}
Li, J.; Lou, X.; Wang, J.; Wang, Y.; Wang, Y.; Xia, S.-T.; and Chen, B. 2026{\natexlab{c}}.
\newblock Imagine Before Concentration: Diffusion-Guided Registers Enhance Partially Relevant Video Retrieval.
\newblock \emph{arXiv preprint arXiv:2604.03653}.

\bibitem[{Li et~al.(2025)Li, Wang, Tan, Lian, Chen, Wang, Zhang, Xia, and Chen}]{hlformer}
Li, J.; Wang, J.; Tan, C.; Lian, N.; Chen, L.; Wang, Y.; Zhang, M.; Xia, S.-T.; and Chen, B. 2025.
\newblock Hlformer: Enhancing partially relevant video retrieval with hyperbolic learning.
\newblock \emph{arXiv preprint arXiv:2507.17402}.

\bibitem[{Luo et~al.(2022)Luo, Ji, Zhong, Chen, Lei, Duan, and Li}]{clip4clip}
Luo, H.; Ji, L.; Zhong, M.; Chen, Y.; Lei, W.; Duan, N.; and Li, T. 2022.
\newblock Clip4clip: An empirical study of clip for end to end video clip retrieval and captioning.
\newblock \emph{Neurocomputing}, 508: 293--304.

\bibitem[{Moon et~al.(2025)Moon, Cho, Jun, Kim, Lee, Wee, Shim, and Heo}]{protoprvr}
Moon, W.; Cho, C.-H.; Jun, W.; Kim, T.; Lee, I.; Wee, D.; Shim, M.; and Heo, J.-P. 2025.
\newblock Prototypes are balanced units for efficient and effective partially relevant video retrieval.
\newblock In \emph{Proceedings of the IEEE/CVF International Conference on Computer Vision}, 21789--21799.

\bibitem[{Moon et~al.(2026)Moon, Jung, Park, Kim, Cho, Jun, and Heo}]{mscprvr}
Moon, W.; Jung, M.; Park, G.; Kim, T.-Y.; Cho, C.-H.; Jun, W.; and Heo, J.-P. 2026.
\newblock Mitigating semantic collapse in partially relevant video retrieval.
\newblock \emph{Advances in Neural Information Processing Systems}, 38: 23196--23217.

\bibitem[{Wang et~al.(2024{\natexlab{a}})Wang, Wang, Chen, Dai, Luo, and Xia}]{gmmformerv2}
Wang, Y.; Wang, J.; Chen, B.; Dai, T.; Luo, R.; and Xia, S.-T. 2024{\natexlab{a}}.
\newblock Gmmformer v2: An uncertainty-aware framework for partially relevant video retrieval.
\newblock \emph{arXiv preprint arXiv:2405.13824}.

\bibitem[{Wang et~al.(2024{\natexlab{b}})Wang, Wang, Chen, Zeng, and Xia}]{gmmformer}
Wang, Y.; Wang, J.; Chen, B.; Zeng, Z.; and Xia, S.-T. 2024{\natexlab{b}}.
\newblock Gmmformer: Gaussian-mixture-model based transformer for efficient partially relevant video retrieval.
\newblock In \emph{Proceedings of the AAAI conference on artificial intelligence}, volume~38, 5767--5775.

\bibitem[{Wu et~al.(2023)Wu, Luo, Fang, Wang, and Ouyang}]{cap4video}
Wu, W.; Luo, H.; Fang, B.; Wang, J.; and Ouyang, W. 2023.
\newblock Cap4video: What can auxiliary captions do for text-video retrieval?
\newblock In \emph{Proceedings of the IEEE/CVF conference on computer vision and pattern recognition}, 10704--10713.

\bibitem[{Yang et~al.(2026)Yang, Wang, Jin, Ma, Xu, and Pang}]{kdcnet}
Yang, J.; Wang, Q.; Jin, Y.; Ma, S.; Xu, M.; and Pang, S. 2026.
\newblock Knowledge-Refined Dual Context-Aware Network for Partially Relevant Video Retrieval.
\newblock \emph{arXiv preprint arXiv:2603.23902}.

\bibitem[{Zhang et~al.(2025{\natexlab{a}})Zhang, Song, Dong, Li, and Yang}]{ral}
Zhang, L.; Song, P.; Dong, J.; Li, K.; and Yang, X. 2025{\natexlab{a}}.
\newblock Enhancing partially relevant video retrieval with robust alignment learning.
\newblock \emph{arXiv preprint arXiv:2509.01383}.

\bibitem[{Zhang et~al.(2025{\natexlab{b}})Zhang, Shao, Chen, Zhang, Zhou, Guan, and Nie}]{falcon}
Zhang, R.; Shao, R.; Chen, G.; Zhang, M.; Zhou, K.; Guan, W.; and Nie, L. 2025{\natexlab{b}}.
\newblock Falcon: Resolving visual redundancy and fragmentation in high-resolution multimodal large language models via visual registers.
\newblock In \emph{Proceedings of the IEEE/CVF International Conference on Computer Vision}, 23530--23540.

\bibitem[{Zhu et~al.(2025)Zhu, Wang, Yang, Yang, Tu, and Wang}]{regqav}
Zhu, X.; Wang, S.; Yang, J.; Yang, Y.; Tu, W.; and Wang, Z. 2025.
\newblock Query-Based Audio-Visual Temporal Forgery Localization with Register-Enhanced Representation Learning.
\newblock In \emph{Proceedings of the 33rd ACM International Conference on Multimedia}, 8547--8556.

\end{thebibliography}

\section*{Appendix}

This Technical Supplement is organised in two parts. Section~A reports additional experiments and diagnostics: hyper-parameter sensitivity, transfer to earlier non-register-based backbones, hard-negative subsets, paired bootstrap, rank-level correction, and routing corruption. Section~B reports method details: full implementation, training schedule, and theoretical properties of the concentration operator.

\subsection{A.1 Hyper-Parameter Sensitivity}

TRACE introduces three score-level hyperparameters: the fusion weight $\lambda$, the routing strength $\gamma$, and the soft register evidence weight $\eta$. In the main experiments, we use a unified configuration across all datasets, i.e., $\lambda=0.03$, $\gamma=0.3$, and $\eta=0.2$. The only dataset-dependent setting is the number of registers $N_r$, which is inherited from the DreamPRVR backbone: $N_r=4$ for ActivityNet Captions, $N_r=6$ for Charades-STA, and $N_r=8$ for TVR. Since changing $N_r$ alters the register generation module rather than only the proposed TRACE calibration branch, we keep this backbone design unchanged.

\begin{figure*}[t]
 \centering
 \includegraphics[width=0.95\linewidth]{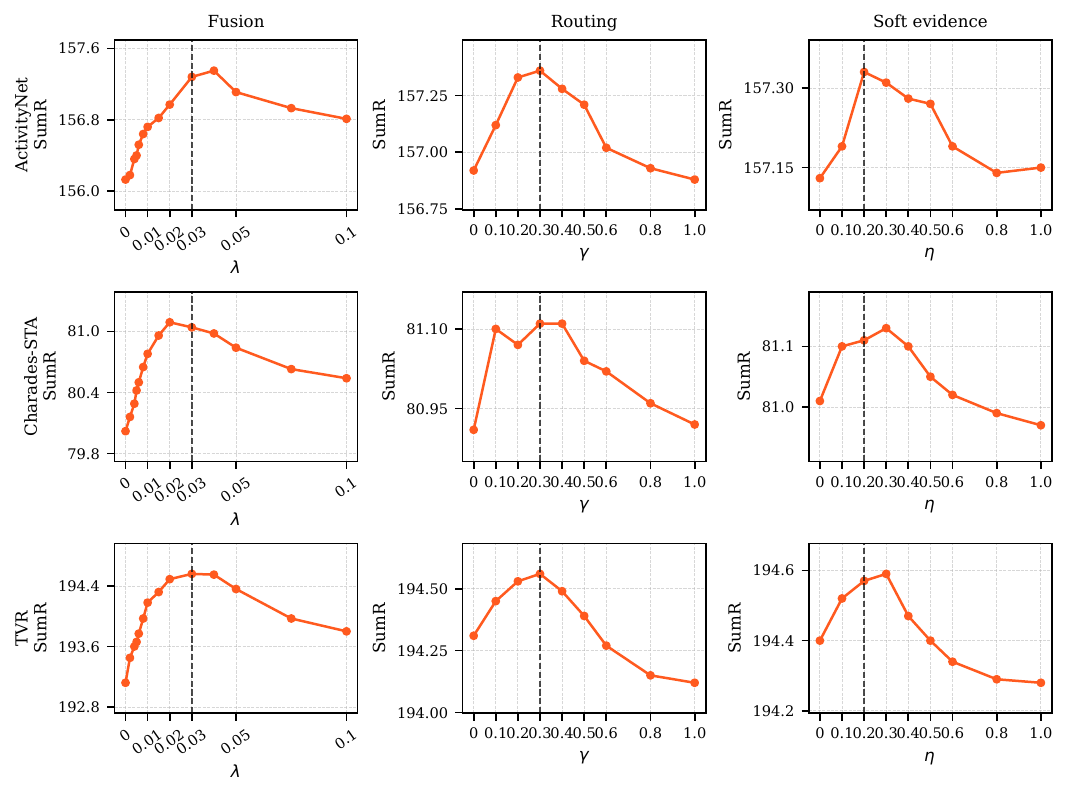}
 \caption{Hyper-parameter analysis of TRACE on validation sets. We evaluate the effects of the score fusion weight $\lambda$, routing strength $\gamma$, and soft evidence weight $\eta$ on SumR. The results show that TRACE is robust within a reasonable range and that a unified score-level configuration works consistently across datasets.}
 \label{fig:hparam_sensitivity}
\end{figure*}

Figure~\ref{fig:hparam_sensitivity} analyzes the influence of $\lambda$, $\gamma$, and $\eta$ on validation SumR. The results show that TRACE maintains stable performance around the unified setting. In particular, non-zero fusion weights consistently outperform $\lambda=0$, showing that text-conditioned evidence provides useful score-level calibration. Moderate routing and soft evidence weights also perform competitively, supporting our choice of a single cross-dataset configuration rather than dataset-specific tuning.

\subsection{A.2 Transfer to Earlier Non-Register-Based PRVR Backbones}

TRACE is instantiated with diffusion-generated registers in the main experiments because they provide explicit global evidence anchors. To test whether the score-time verification principle is strictly tied to such registers, we transfer the same idea to earlier non-register-based PRVR backbones. For each method and dataset, we use the corresponding official checkpoint, keep the backbone frozen, and derive lightweight anchors from its encoded local tokens by temporal partitioning or global pooling. Candidate anchor constructions and fusion weights are selected using validation SumR only; the selected configuration is then fixed before evaluation. We apply the resulting query-anchor-frame evidence route to calibrate the original similarity. This diagnostic does not retrain the baselines and should be interpreted as an anchor-agnostic scoring test rather than a full backbone-level reimplementation.

The anchor construction follows the representation structure of each backbone. MS-SL builds multi-scale local matching features, so we construct anchors from its encoded frame tokens. GMMFormer models temporal evidence mainly through Gaussian clip-level representations; accordingly, we use clip-token anchors for GMMFormer. GMMFormerV2 provides stronger frame-level local tokens, so we derive frame-token anchors. HLFormer models hierarchical partial relevance in hyperbolic space while still producing local evidence for final ranking; we therefore construct anchors from its encoded frame tokens.

\begin{table*}[t]
\centering
\small
\setlength{\tabcolsep}{4pt}
\begin{tabular}{llccclc}
\toprule
Backbone & Dataset & Base SumR & +Route SumR & $\Delta$SumR & Selected Anchor & $\lambda$ \\
\midrule
MS-SL & ActivityNet & 140.10 & 140.28 & +0.18 & global-4 & 0.015 \\
MS-SL & Charades & 68.40 & 68.53 & +0.13 & temporal-4 & 0.010 \\
MS-SL & TVR & 172.40 & 172.61 & +0.21 & temporal-6 & 0.010 \\
GMMFormer & ActivityNet & 146.00 & 146.24 & +0.24 & temporal-4 & 0.060 \\
GMMFormer & Charades & 72.90 & 73.37 & +0.47 & temporal-6 & 0.120 \\
GMMFormer & TVR & 176.60 & 177.10 & +0.50 & temporal-8 & 0.200 \\
GMMFormerV2 & ActivityNet & 154.90 & 155.44 & +0.54 & temporal-8 & 0.200 \\
GMMFormerV2 & Charades & 78.20 & 78.51 & +0.31 & temporal-6 & 0.100 \\
GMMFormerV2 & TVR & 189.10 & 189.98 & +0.88 & temporal-4 & 0.200 \\
HLFormer & ActivityNet & 154.90 & 155.79 & +0.89 & temporal-6 & 0.150 \\
HLFormer & Charades & 78.70 & 79.26 & +0.56 & temporal-6 & 0.120 \\
HLFormer & TVR & 187.70 & 189.13 & +1.43 & global-4 & 0.200 \\
\bottomrule
\end{tabular}
\caption{Full frozen-backbone diagnostic on earlier non-register-based PRVR backbones. Each row uses the official dataset-specific checkpoint of the corresponding method. Anchor type and $\lambda$ are selected on the validation split and then frozen for evaluation; all backbone parameters remain frozen.}
\label{tab:non_register_appendix}
\end{table*}

Table~\ref{tab:non_register_appendix} shows that anchor-based score verification yields positive gains across the evaluated earlier PRVR backbones. The gains are smaller than the main TRACE results with explicit diffusion registers, which is expected because these anchors are obtained by simple pooling rather than learned global register generation. Nevertheless, the diagnostic supports our interpretation that the key operation is score-time verification through query-selected global anchors, while diffusion registers are an effective instantiation rather than the only possible source of anchors. We also attempted to include CLIP4Clip, Cap4Video, XML, and CONQUER in this diagnostic. Their released code or checkpoints do not directly provide comparable three-dataset frozen-backbone evaluation with exposed local evidence tokens, so we keep them as standard comparison baselines in Table~1 rather than reporting non-comparable diagnostic results.

\subsection{A.3 Additional Diagnostics}

\begin{figure*}[t]
 \centering
 \includegraphics[width=0.95\linewidth]{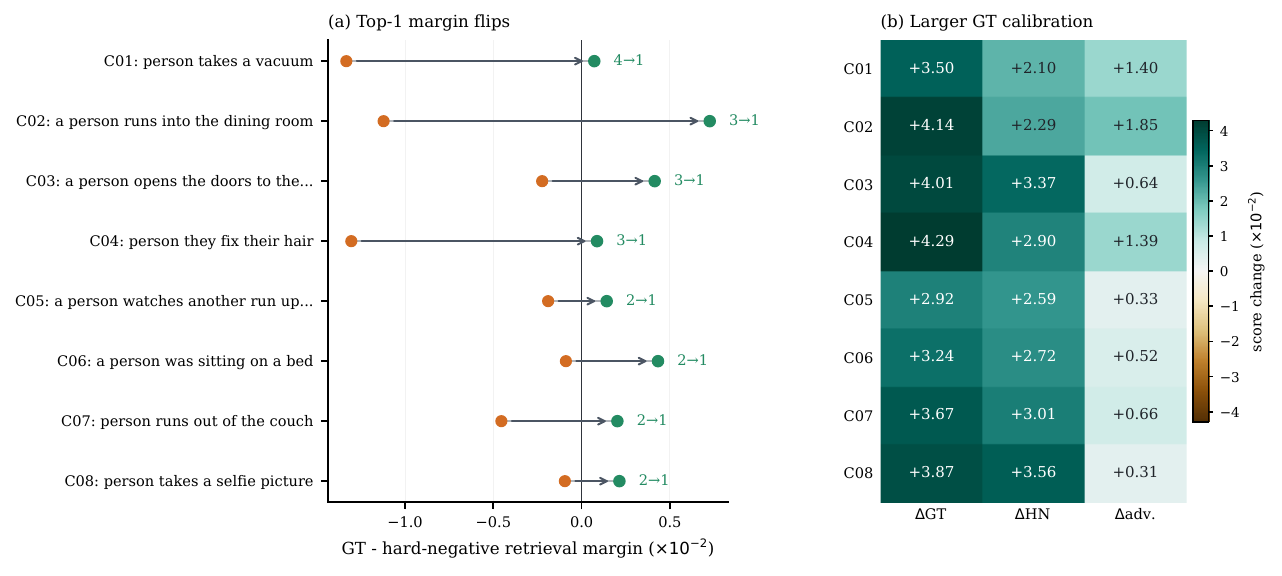}
 \caption{Top-rank correction on Charades-STA hard-negative cases. Left: for each query, the baseline ranks a hard negative above the ground truth, and TRACE moves the margin above zero to restore rank 1. Right: the score-change decomposition into $\Delta$GT, $\Delta$HN, and the relative advantage $\Delta$adv$=\Delta$GT$-\Delta$HN.}
 \label{fig:trace_case_story}
\end{figure*}

We first provide additional qualitative visualizations that complement the main-text result figure. The main qualitative result already illustrates a corrected hard-negative case at the frame-evidence level, showing that the ground-truth peak is better aligned with register support than the hard-negative peak. Figure~\ref{fig:trace_case_story} further summarizes strict top-1 correction cases, where TRACE consistently gives the ground-truth video a larger calibration boost than the hard negative.

We next evaluate whether TRACE is especially helpful on the failure mode it is designed to address. We first identify queries for which the baseline ranks an incorrect video at top 1. For each such query, we measure the unsupported local peak of that top-1 hard negative, i.e., the gap between its strongest direct local evidence and the register support at the same peak. We then rank these baseline-error queries by the unsupported gap and evaluate TRACE on the top 50\% and top 25\% subsets. If TRACE only introduced a generic score bias, its gain would not systematically grow on these increasingly difficult subsets.

\begin{table*}[t]
\centering
\small
\setlength{\tabcolsep}{5pt}
\begin{tabular}{lrrrrrr}
\toprule
Dataset & All $\Delta$SumR & Err. $\Delta$SumR & Top-50 $\Delta$SumR & Top-25 $\Delta$SumR & Top-25 Mean Gain & Top-25 \#Query \\
\midrule
ActivityNet & +1.20 & +1.48 & +1.70 & \textbf{+2.08} & +1.96 & 3584 \\
Charades-STA & +1.10 & +1.30 & +1.48 & \textbf{+1.82} & +1.85 & 906 \\
TVR & +1.50 & +1.78 & +1.96 & \textbf{+2.34} & +2.48 & 2279 \\
\bottomrule
\end{tabular}
\caption{Hard-negative subset analysis on evaluation queries. Err. denotes queries for which the baseline top-1 video is incorrect. Top-50 and Top-25 are the 50\% and 25\% subsets of these baseline-error queries with the largest unsupported local gaps. $\Delta$SumR is the recall improvement on each subset; Mean Gain is the average ground-truth rank improvement on Top-25.}
\label{tab:hard_negative_subset}
\end{table*}

Table~\ref{tab:hard_negative_subset} shows that TRACE's gains become larger on hard-negative subsets across all three datasets. The all-query gains match the DreamPRVR-to-TRACE SumR differences in Table~1, while the gains further increase on baseline-error queries and on the top 25\% unsupported hard-negative subset. This subset trend is important because it connects the numerical gains to the proposed failure mode: TRACE is particularly effective when the baseline is distracted by locally plausible but globally weakly supported negative evidence.

For statistical testing, we treat each query as one paired observation and retain the pairing between the base and TRACE ranks. We draw 20,000 bootstrap samples of query indices with replacement and report the 2.5th and 97.5th percentiles of the resulting $\Delta$SumR distribution as the 95\% confidence interval. The reported $p$-values are obtained from a one-sided paired sign-flip test with 20,000 random sign assignments. All tests use the same evaluation split as the corresponding diagnostic, with random seed 20260712.

\begin{table*}[t]
\centering
\small
\setlength{\tabcolsep}{4pt}
\begin{tabular}{lccccl}
\toprule
Dataset & All $\Delta$SumR & 95\% CI & $p$ & Top-25 $\Delta$SumR & Top-25 95\% CI / $p$ \\
\midrule
ActivityNet & +1.20 & [+0.84, +1.56] & $<0.001$ & +2.08 & [+1.36, +2.82] / $<0.001$ \\
Charades-STA & +1.10 & [+0.52, +1.69] & $<0.001$ & +1.82 & [+0.58, +3.16] / 0.006 \\
TVR & +1.50 & [+1.16, +1.84] & $<0.001$ & +2.34 & [+1.55, +3.20] / $<0.001$ \\
\bottomrule
\end{tabular}
\caption{Query-level paired bootstrap confidence intervals and paired sign-flip tests for SumR improvements. The all-query interval measures the recall improvement over the full evaluation split. The Top-25 columns repeat the analysis on the 25\% baseline-error queries whose top-1 hard negatives have the largest unsupported local peaks.}
\label{tab:paired_significance}
\end{table*}

Table~\ref{tab:paired_significance} further quantifies this trend with paired bootstrap confidence intervals. The full-query SumR improvement is statistically reliable across all three datasets, and the top-25\% unsupported hard-negative subsets show even larger positive effects. This supports our interpretation that TRACE improves the overall retrieval score while being especially effective on the intended hard-negative boundary.

\begin{figure*}[t]
 \centering
 \includegraphics[width=0.95\linewidth]{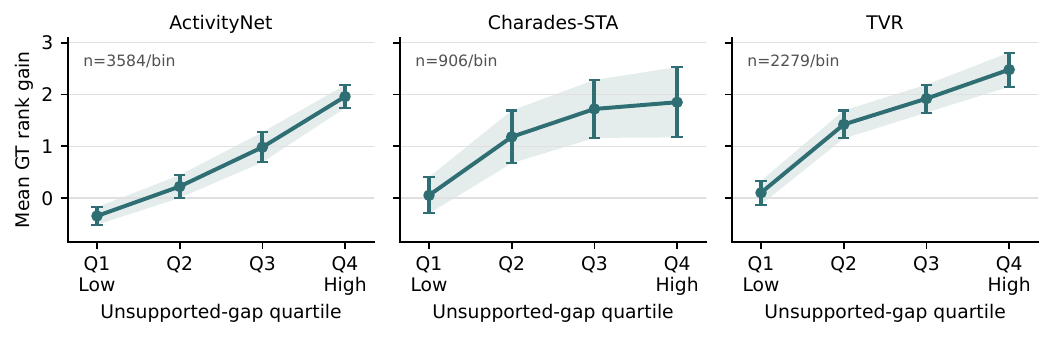}
 \caption{Unsupported-gap stratification on baseline-error evaluation queries. Queries are split into four quartile bins (Q1--Q4) by the unsupported evidence gap between the baseline top-1 hard negative and the ground truth. Larger gaps yield larger rank gains after TRACE, and Q4 aligns with the Top-25 subset in Table~\ref{tab:hard_negative_subset}.}
 \label{fig:unsupported_rank_gain}
\end{figure*}

\begin{table*}[t]
\centering
\small
\setlength{\tabcolsep}{5pt}
\begin{tabular}{lrrrrrrr}
\toprule
Dataset & Base R@1 & TRACE R@1 & $\Delta$R@1 & Rank$\uparrow$ & Rank$\downarrow$ & Mean Gain & Median Gain \\
\midrule
ActivityNet & 8.7 & 9.0 & +0.3 & 5204 & 2841 & +1.28 & 0.00 \\
Charades-STA & 2.6 & 2.7 & +0.1 & 1938 & 1096 & +1.55 & 0.00 \\
TVR & 17.4 & 17.8 & +0.4 & 3518 & 1292 & +1.76 & 0.00 \\
\bottomrule
\end{tabular}
\caption{Rank-level change statistics on evaluation queries. Base and TRACE R@1 repeat the values in Table~1, making the top-rank change directly comparable to the main results. Rank$\uparrow$ and Rank$\downarrow$ count queries whose ground-truth video rank improves or degrades, while Mean Gain and Median Gain summarize the signed rank improvement.}
\label{tab:rank_correction}
\end{table*}

To further test whether TRACE benefits from meaningful cross-granular grounding rather than merely adding a generic score bias, we conduct a routing corruption diagnostic on TVR. We compare the base branch, the original TRACE score, and three corrupted routing variants. Uniform $q{\to}R$ removes query-specific register selection, cross-video shuffle $q{\to}R$ replaces the activation with mismatched query-register correspondence, while shuffle $R{\to}F$ corrupts the register-to-frame grounding path before computing routed evidence. These corruptions preserve the existence of an additional score branch but progressively destroy the semantic correspondence of its route.

\begin{table*}[t]
\centering
\small
\setlength{\tabcolsep}{4pt}
\begin{tabular}{lccccc|rrrr}
\toprule
Variant & R@1 & R@5 & R@10 & R@100 & SumR
& $\Delta$SumR & Rank$\uparrow$ & Rank$\downarrow$ & Mean Gain \\
\midrule
Base branch & 17.4 & 39.0 & 50.4 & 86.2 & 193.1
& -- & -- & -- & -- \\
TRACE & \textbf{17.8} & \textbf{39.4} & \textbf{50.7} & \textbf{86.7} & \textbf{194.6}
& \textbf{+1.5} & \textbf{3518} & 1292 & \textbf{+1.76} \\
Uniform $q{\to}R$ & 17.5 & 39.2 & 50.6 & 86.5 & 193.9
& +0.8 & 2864 & 1847 & +0.74 \\
Cross-video shuffle $q{\to}R$ & 17.4 & 39.1 & 50.5 & 86.4 & 193.4
& +0.3 & 2416 & 2189 & +0.37 \\
Shuffle $R{\to}F$ & 17.2 & 38.9 & 50.2 & 86.3 & 192.6
& -0.5 & 2527 & 3725 & -0.24 \\
\bottomrule
\end{tabular}
\caption{Routing corruption diagnostics on TVR. $\Delta$SumR is measured relative to the base branch. Rank$\uparrow$ and Rank$\downarrow$ count queries whose ground-truth video rank is improved or degraded, respectively. Uniform $q{\to}R$ removes query-specific register selection, cross-video shuffling further breaks the query-register correspondence, and shuffling $R{\to}F$ corrupts the grounding from registers to frame evidence.}
\label{tab:routing_corruption}
\end{table*}

Table~\ref{tab:routing_corruption} shows that TRACE improves the base branch across multiple recall cutoffs, increasing SumR from 193.1 to 194.6 and improving the ground-truth rank for 3,518 TVR queries. This indicates that the effect is not limited to strict top-1 boundary cases. Removing query-specific activation or replacing it with cross-video activation weakens the gain, while corrupting the register-to-frame grounding path reduces SumR to 192.6 and degrades more ranks than it improves. This diagnostic supports the role of grounded $q{\to}R{\to}F$ evidence in score-time verification rather than a generic score offset.

\subsection{A.4 Scope and Limitations}

TRACE studies score-time evidence verification rather than replacing the underlying video-language backbone. Its current instantiation uses global registers because they provide explicit semantic anchors that can be selected by the query and grounded to frames. Therefore, the method is most directly applicable to PRVR systems that expose both local temporal tokens and global evidence anchors. This scope also explains why the observed gains are moderate but consistent: TRACE preserves the base representation space and focuses on correcting ranking behavior through text-conditioned evidence verification. Importantly, this limitation is also a design choice: by operating at the score level, TRACE can diagnose whether a local peak is globally supported without retraining a heavy generator or altering the backbone architecture.

We organize the remaining discussion along four explicit axes: register capacity design, explicit failure modes, the high-precision--mid-recall trade-off, and the computational budget.

\paragraph{Register capacity $N_r$.}
The number of registers is inherited from the DreamPRVR backbone ($N_r{=}4$ for ActivityNet Captions, $N_r{=}6$ for Charades-STA, $N_r{=}8$ for TVR) rather than treated as a TRACE-specific hyperparameter, because changing $N_r$ modifies the register generator itself and therefore the representation space shared with the base branch. The dataset-specific choice reflects an empirical balance between query vocabulary breadth and per-register specialization. ActivityNet Captions queries tend to describe a single salient activity, so a small register set is sufficient; Charades-STA queries mix action verbs with multiple fine-grained entities, which benefits from a moderately larger set; TVR queries describe character interactions and scene-level relations across longer videos, which motivates the largest set. We did not sweep $N_r$ independently because such a sweep would change the backbone's representation capacity rather than the score-time rule, and a register-agnostic verification result is the very claim of Table~\ref{tab:non_register_appendix}.

\paragraph{Explicit failure modes.}
Two concrete regimes degrade TRACE back toward its base branch. First, when a query is semantically distant from every register in the active video, the softmax in Eq.~(4) becomes approximately uniform and $\rho_i$ degrades to a query-agnostic global support term. This regime corresponds to the \emph{Uniform $q{\to}R$} corruption diagnostic in Table~\ref{tab:routing_corruption}, where TVR $\Delta$SumR drops from $+1.5$ to $+0.8$ and the number of rank-improved queries drops from $3{,}518$ to $2{,}864$. The remedy is therefore not a fix inside TRACE but a richer register vocabulary, e.g.\ a hybrid register set that mixes global, local, and event-level anchors. Second, TRACE relies on query-anchor affinity being meaningful for the current ranking. When explicit learned registers are unavailable and anchors must instead be derived by simple pooling from a frozen non-register backbone, the gain falls into the lightweight-anchor range of roughly $+0.13$ to $+1.43$ SumR (Table~\ref{tab:non_register_appendix}), with the strongest lightweight-anchor gain ($+1.43$ on HLFormer TVR) approaching but not matching the diffusion-register instantiation.

\paragraph{High-precision vs.\ mid-recall trade-off.}
Although TRACE achieves the best SumR on all three datasets, it does not dominate every individual recall cutoff. The pattern is consistent across datasets: TRACE improves R@1 (high-precision boundary) and R@100 (long-tail boundary) but moves slightly at R@5 or R@10 on certain datasets, e.g.\ Charades-STA R@10 changes from $14.5$ to $14.3$ (Table~1). This indicates that score-time verification and uncertainty-aware evidence estimation emphasize complementary aspects of the ranking distribution: routed support is most informative when the top of the list and the long tail are the contested regions, whereas mid-recall cutoffs are dominated by relative ordering among already-strong candidates, where routed support contributes a smaller per-item margin. The trade-off is therefore not a flaw but a profile of what score-time evidence verification can and cannot move.

\paragraph{Computational budget.}
TRACE adds a small score-time branch without introducing an additional backbone pass, so the dominant inference cost remains the base forward pass. The additional cost per query comes from (i) a register activation vector $\pi\in\mathbb{R}^{N_r}$ computed as one $N_r$-way softmax over cosine similarities, (ii) an $N_r \times F$ register--frame compatibility matrix computed for evidence routing, and (iii) a per-frame reduction that takes the temporal maximum of $d_i + \gamma\rho_i$. Thus, the dominant added cost is the $N_r \times F$ register--frame aggregation and a single temporal max-reduction; the memory footprint is proportional to $N_r \times F$ per video. Since $N_r$ is small, the verification cost remains bounded by the base forward pass as the video encoder scales. A natural next step is to combine TRACE-style routed support with an evidential head, but we leave that direction for future work.

Finally, the main-text results use diffusion-generated registers because they consistently provide the strongest global anchors. Table~\ref{tab:non_register_appendix} reports the corresponding lightweight-anchor sweep, where global pooling and temporal partitioning replace the register generator while leaving the rest of the pipeline intact. Even with simple anchors, the verification branch still yields positive $\Delta$SumR across all four backbones and all three datasets, supporting score-time verification as the shared operation. The much larger gains on TVR (up to $+1.43$ $\Delta$SumR on HLFormer) further suggest that richer, query-relevant anchor generation is an important source of TRACE's headline numbers, while the verification rule itself is anchor-agnostic.

\subsection{B.1 Implementation and Computation Details}

For ActivityNet Captions and Charades-STA, we use the provided I3D features as video representations. Query representations are encoded with 1,024-dimensional RoBERTa features following MS-SL. For TVR, we use the released 3,072-dimensional video features that concatenate frame-level ResNet152 and segment-level I3D representations, and encode the corresponding textual queries with 768-dimensional RoBERTa features.

We follow the register-based backbone configuration and use a hidden dimension of 384 with 4 attention heads. The number of registers is set to 4, 6, and 8 for ActivityNet Captions, Charades-STA, and TVR, respectively, directly following the DreamPRVR backbone design. Since changing $N_r$ modifies the register generator itself, we do not treat it as a TRACE-specific hyperparameter. For the proposed score-level branch, we use the same configuration across all datasets: the evidence fusion weight is $\lambda=0.03$, the routing strength is $\gamma=0.3$, and the soft register evidence weight is $\eta=0.2$. The register activation temperature $\tau_r$ is fixed to 0.07. The TRACE auxiliary objective uses $\lambda_{\mathrm{nce}}=0.02$ and $\lambda_{\mathrm{tri}}=0.05$, with triplet margins 0.2 for ActivityNet Captions and Charades-STA and 0.1 for TVR.

\begin{algorithm}[t]
\caption{Text-Conditioned Register Activation and Cross-Granular Evidence Routing (TRACE)}
\label{alg:trace}
\begin{algorithmic}[1]
\REQUIRE Query embedding $q$, frame tokens $F$, clip tokens $C$ (consumed by $S_{\mathrm{base}}$ via Eq.~(1)), and video registers $R$
\STATE Compute base similarity $S_{\mathrm{base}}$ by Eq.~(1).
\STATE Compute query-to-frame evidence $d_i=\cos(q,f_i)$.
\STATE Compute query-register affinity $a_k=\cos(q,r_k)$.
\STATE Compute register-frame compatibility $b_{k,i}=\cos(r_k,f_i)$.
\STATE Route text-selected registers to frames to obtain $\rho_i$.
\STATE Combine local and routed evidence: $\tilde{d}_i=d_i+\gamma\rho_i$.
\STATE Select the strongest routed evidence $S_{\mathrm{route}}=\max_i\tilde{d}_i$.
\STATE Compute register-guided soft evidence $S_{\mathrm{soft}}$.
\STATE Compute $S_{\mathrm{trace}}=S_{\mathrm{route}}+\eta S_{\mathrm{soft}}$.
\RETURN $S_{\mathrm{final}}=S_{\mathrm{base}}+\lambda S_{\mathrm{trace}}$.
\end{algorithmic}
\end{algorithm}

Training follows one continuous progressive procedure. We first warm up the base register-based backbone for 100 epochs with $\mathcal{L}_{\mathrm{base}}$ and use its resulting retrieval score as $S_{\mathrm{base}}$; all epoch indices $t$ below are local to the subsequent TRACE phase. During the first 5 TRACE-stage epochs ($t=0,\ldots,4$), we freeze the backbone and optimize the evidence projections with $\mathcal{L}_{\mathrm{trace}}$, while the ranking score remains $S_{\mathrm{base}}$ because the effective integration weight is zero. From TRACE-stage epoch $t=5$, we jointly optimize the backbone and evidence branch with $\mathcal{L}_{\mathrm{base}}+\mathcal{L}_{\mathrm{trace}}$. From $t=30$, score integration is activated and the effective weight is linearly ramped for 20 epochs:
\begin{equation}
\lambda(t)=\lambda\cdot\min\left(1,\frac{t-30}{20}\right), \quad t\ge 30 .
\end{equation}
The final calibration learning rates are $8\times10^{-5}$, $5\times10^{-5}$, and $2\times10^{-6}$ for ActivityNet Captions, Charades-STA, and TVR, respectively. All models are implemented in PyTorch and trained with AdamW using $\beta_1{=}0.9$, $\beta_2{=}0.98$, $\epsilon{=}10^{-8}$, and a weight decay of $10^{-4}$ applied to all parameters except layer-norm scales and bias terms. We use a linear warm-up over the first $5$ TRACE-stage epochs followed by a cosine decay to $10\%$ of the peak learning rate over the remaining epochs, and a gradient clip of $1.0$ applied to the global norm. The mini-batch size is $128$ on eight NVIDIA A800-SXM4-80GB GPUs using bfloat16 mixed-precision training. The software environment is Ubuntu 22.04.5 LTS with Python 3.10.18, PyTorch 2.6.0+cu126, CUDA 12.6, and NVIDIA driver 550.127.08. We follow the standard preprocessing inherited from each backbone and do not introduce additional data augmentation in TRACE. Each reported experiment is run three times with independent seeds, and the released training configuration records the corresponding seed settings. Final evaluation checkpoints are obtained by snapshot averaging the last three epochs of TRACE-stage training, a common practice that reduces single-seed variance relative to last-epoch-only reporting without changing the mean. We will release code, pretrained checkpoints, and evaluation scripts upon acceptance.

\subsection{B.2 Theoretical Properties of Hierarchical Evidence Concentration}

We analyze the core concentration operator to clarify its relation to standard PRVR scoring and its ability to correct spurious local peaks. Throughout this section, cosine similarities such as $\cos(q,f_i)$ are treated as unnormalized energy scores under the energy-based interpretation of Eq.~(3), not as proper probabilities; this convention follows the standard practice in PRVR analysis and keeps the notation aligned with the main text. We first collect the symbols used below, then state the setup and the main correction result.

\begin{table}[h]
\centering
\small
\resizebox{\columnwidth}{!}{
\begin{tabular}{ll}
\toprule
Symbol & Meaning \\
\midrule
$q$ & query embedding (single vector) \\
$F=\{f_1,\ldots,f_F\}$ & per-frame visual token sequence of length $F$ \\
$R=\{r_1,\ldots,r_{N_r}\}$ & global register set of size $N_r$ \\
$d_i=\cos(q,f_i)$ & direct query--frame evidence at frame $i$ \\
$a_k=\cos(q,r_k)$ & query--register affinity at register $k$ \\
$b_{k,i}=\cos(r_k,f_i)$ & register--frame compatibility at register $k$, frame $i$ \\
$\tau_r>0$ & register activation temperature \\
$\pi_k=\exp(a_k/\tau_r)/\sum_{k'}\exp(a_{k'}/\tau_r)$ & query-conditioned register weight \\
$\rho_i=\tau_r\log\sum_k\exp\!\left((a_k+b_{k,i})/\tau_r\right)$ & routed support at frame $i$ \\
$\gamma>0$ & routing strength in $S_{\mathrm{route}}$ \\
\bottomrule
\end{tabular}}
\caption{Notation used in Technical Supplement, Sec.~B.2. All symbols match the main-text definitions (Eqs.~(4)--(6)), with cosine similarities treated as energy scores rather than probabilities.}
\label{tab:notation}
\end{table}

\paragraph{Setup.}
We consider a single query $q$ and a single video $V$ with frame tokens $F=\{f_1,\ldots,f_F\}$ and global registers $R=\{r_1,\ldots,r_{N_r}\}$ (Table~\ref{tab:notation} collects all symbols used below). The standard PRVR local selector is $S_{\max}(q,V)=\max_i d_i$ with $d_i=\cos(q,f_i)$, and the TRACE routed selector is $S_{\mathrm{route}}(q,V)=\max_i\!\left[d_i+\gamma\rho_i\right]$ with $\rho_i=\tau_r\log\sum_{k=1}^{N_r}\exp\!\left((a_k+b_{k,i})/\tau_r\right)$, where $a_k=\cos(q,r_k)$ and $b_{k,i}=\cos(r_k,f_i)$. We make the following mild assumptions: (i) all cosine similarities lie in $[-1,1]$, so $d_i, a_k, b_{k,i}$ are bounded; (ii) the selected frame indices $i^+$ and $i^-$ in $V^+$ and $V^-$ are those that achieve the inner maximum, which is well-defined because $F$ is finite; and (iii) the routing strength $\gamma$ and temperature $\tau_r$ are positive hyperparameters chosen once and held fixed during inference. No assumption is required on the distribution of $d_i$, $a_k$, or $b_{k,i}$ beyond boundedness. Standard max-style PRVR scoring can be written as
\begin{equation}
S_{\max}(q,V)=\max_i d_i .
\end{equation}
This form is suitable for partial relevance because only one moment may match the query. However, it is also a degenerate evidence selector: the selected frame only needs to maximize local query similarity, without being verified by any query-relevant global evidence. Thus, a negative video with one coincidentally similar local fragment can dominate the final ranking.

TRACE changes this selection rule by adding text-conditioned global support. Recall that the routed support for frame $i$ is
\begin{equation}
\rho_i=\tau_r\log\sum_{k=1}^{N_r}\exp\left((a_k+b_{k,i})/\tau_r\right), \label{eq:appendix_lse_route}
\end{equation}
where $a_k=\cos(q,r_k)$ measures query-register affinity and $b_{k,i}=\cos(r_k,f_i)$ measures register-frame compatibility. For $x_k=a_k+b_{k,i}$, log-sum-exp satisfies
\begin{equation}
\max_k x_k\leq\tau_r\log\sum_k\exp(x_k/\tau_r)\leq\max_k x_k+\tau_r\log N_r, \label{eq:appendix_lse_bound}
\end{equation}
Thus, register-path aggregation is a bounded smooth approximation to selecting the strongest latent $q{\to}R{\to}F$ path. The outer temporal maximum then gives the hierarchical operator:
\begin{equation}
S_{\mathrm{route}}(q,V)=\max_i\left[d_i+\gamma\rho_i\right], \label{eq:appendix_support_regularized}
\end{equation}
Compared with raw max pooling, Eq.~\ref{eq:appendix_support_regularized} selects a local frame only after considering whether there exists a query-relevant global anchor that also supports this frame, while retaining a smooth optimization surface over latent register choices. The approximation gap at each frame is bounded by $\tau_r\log N_r$ and the corresponding routed-score gap is bounded by $\gamma\tau_r\log N_r$.

The operator exposes several meaningful reductions. When $\gamma=0$, Eq.~\ref{eq:appendix_support_regularized} exactly reduces to standard max-style PRVR scoring. When the activation scores $a_k$ are query-independent, the inner aggregation becomes query-agnostic global support, corresponding to the uniform-activation diagnostic in Table~3. When $b_{k,i}$ is removed, register evidence no longer distinguishes temporal locations and reduces to a global query-register bias. Finally, replacing the outer maximum with a temporal average yields the diluted-evidence variant in Table~3. These cases isolate the three defining operations of TRACE: query-conditioned path selection, register-to-frame grounding, and localized temporal concentration.

\paragraph{Theorem 1 (Pairwise correction condition).}
Consider a positive video $V^+$ and a hard negative video $V^-$. Let $d^+$ and $d^-$ be their strongest direct local evidence scores, and let $\rho^+$ and $\rho^-$ be the corresponding routed support values at these selected local peaks. Suppose the raw local selector ranks the hard negative above the positive video, i.e.,
\begin{equation}
d^- > d^+ .
\end{equation}
For this pair of selected peaks, the routed selector reverses their pairwise order whenever
\begin{equation}
\gamma(\rho^+-\rho^-)>d^- - d^+ . \label{eq:appendix_correction_condition}
\end{equation}

\paragraph{Proof.}
Under the routed selector, the positive video is ranked above the hard negative if
\begin{equation}
d^+ + \gamma\rho^+ > d^- + \gamma\rho^- .
\end{equation}
Subtracting $d^-$ from both sides and rearranging gives
\begin{equation}
\gamma(\rho^+-\rho^-) > d^- - d^+ ,
\end{equation}
which is exactly Eq.~\ref{eq:appendix_correction_condition}. $\square$

This condition directly explains the intended top-rank correction behavior. A hard negative may win under $S_{\max}$ because $d^-$ is slightly larger than $d^+$. TRACE can recover the positive video in this pairwise comparison if the local evidence in $V^+$ is better supported by query-activated global registers, i.e., $\rho^+>\rho^-$. Conversely, if a hard negative has both stronger local evidence and comparable or stronger routed support, TRACE should not force a correction. This is why TRACE is best understood as targeted evidence verification rather than a uniform score shift.

\end{document}